%% file: main.tex
\documentclass[runningheads]{llncs}

\usepackage{eccv}

\usepackage{eccvabbrv}

\usepackage{graphicx}
\usepackage{booktabs}
\usepackage{multirow}
\usepackage[accsupp]{axessibility}  

\usepackage[pagebackref,breaklinks,colorlinks,citecolor=eccvblue]{hyperref}

\usepackage{orcidlink}

\begin{document}

\title{Scene-aware Human Motion Forecasting via Mutual Distance Prediction} 


\author{Chaoyue Xing\inst{1} \and
Wei Mao\inst{1} \and
Miaomiao Liu\inst{1}}

\authorrunning{C.~Xing et al.}

\institute{Australian National University, Canberra, Australia \inst{1}
}
\newcommand{\cy}[1]{\textcolor{magenta}{#1}}
\newcommand{\wm}[1]{\textcolor{blue}{#1}}
\maketitle

\input{01-abstract}
\input{02-introduction}
\input{03-relatedwork}
\input{04-methodology}

\input{05-expriments}
\input{06-conclusion}

%
%
\bibliographystyle{splncs04}
\bibliography{main}
\end{document}

%% file: 01-abstract.tex
\begin{abstract}
In this paper, we address the issue of scene-aware 3D human 
motion forecasting. A key challenge in this task is to predict future human motions that are coherent with the scene by modeling human-scene interactions. ~While recent works have demonstrated that explicit constraints on human-scene interactions can prevent the occurrence of~\emph{ghost motion},
they only provide constraints on partial human motion 
e.g., the global motion of the human or a few joints contacting the scene,
leaving the rest of unconstrained.
To address this limitation, we propose to represent the human-scene interaction using the mutual distance between the human body and the scene. Such mutual distances constrain both the local and global human motion, resulting in a \emph{whole-body motion} constrained prediction. In particular, mutual distance constraints
consist of two components, the signed distance of each vertex on the human mesh to the scene surface and the distance of basis scene points to the human mesh. We further introduce a global scene representation learned from a signed distance function (SDF) volume to ensure coherence between the global scene representation and the explicit constraint from the mutual distance. We develop a pipeline with two sequential steps: predicting the future mutual distances first, followed by forecasting future human motion. We explicitly ensure consistency between predicted poses and mutual distances during training. Extensive testing on both synthetic and real datasets demonstrates that our method consistently surpasses the performance of current state-of-the-art techniques.
  \keywords{Scene-aware \and Human Motion Forecasting \and Mutual Distance\and SDF Volume}
\end{abstract}

%% file: 02-introduction.tex
\section{Introduction}\label{sec:intro}
Human motion forecasting has numerous applications, including autonomous driving~\cite{paden2016survey}, human-robot interaction~\cite{koppula2013anticipating}, animation~\cite{van2010real} and virtual/augmented reality~\cite{starke2019neural}. Despite being an active research field for decades~\cite{brand2000style,sidenbladh2002implicit,taylor2007modeling,wang2008gaussian}, most previous methods~\cite{Li_2018_CVPR,aksan2019structured,mao2019learning,wang2019imitation,gopalakrishnan2019neural,li2020dynamic,mao2020history,cai2020learning} focus solely on human motion and overlook the fact that humans interact with the 3D environment. In contrast, this paper addresses the problem of scene-aware human motion prediction.

Recent advancements in scene-aware human synthesis~\cite{wang2021synthesizing,zhang2020place,hassan2021populating} and human motion prediction~\cite{cao2020long,corona2020context,zheng2022gimo,mao2022contact} have begun to model the human-scene interaction. Some approaches achieve this implicitly by incorporating scene encoding as an additional input to their model e.g., 2D scene image~\cite{cao2020long}, 3D scene point cloud~\cite{wang2021synthesizing} or a specific object~\cite{corona2020context}. However, without explicit constraints, those methods tend to predict motions with artifacts e.g., ``ghost motion''. Alternatively, other works~\cite{zhang2020place,zheng2022gimo,mao2022contact, scofano2023staged} start to explore the explicit representation of human-scene interaction. However, none of their representations are able to constrain the \emph{whole-body} motion both locally and globally. For example, the proximity between human and scene used in~\cite{zhang2020place} mainly constrains the global position of the human, while the human gaze employed in~\cite{zheng2022gimo} only specifies the potential target location of the human. Although the contact map proposed in~\cite{mao2022contact, scofano2023staged} achieves detailed constraints on both the local pose and global position of certain human joints, their contact map can only constrain joints that contact the scene, neglecting the others, such as elbow joints, which often remain unsupported.


\begin{figure*}[!ht]
    \centering
    \begin{tabular}{cc}
          \includegraphics[width=0.31\linewidth]{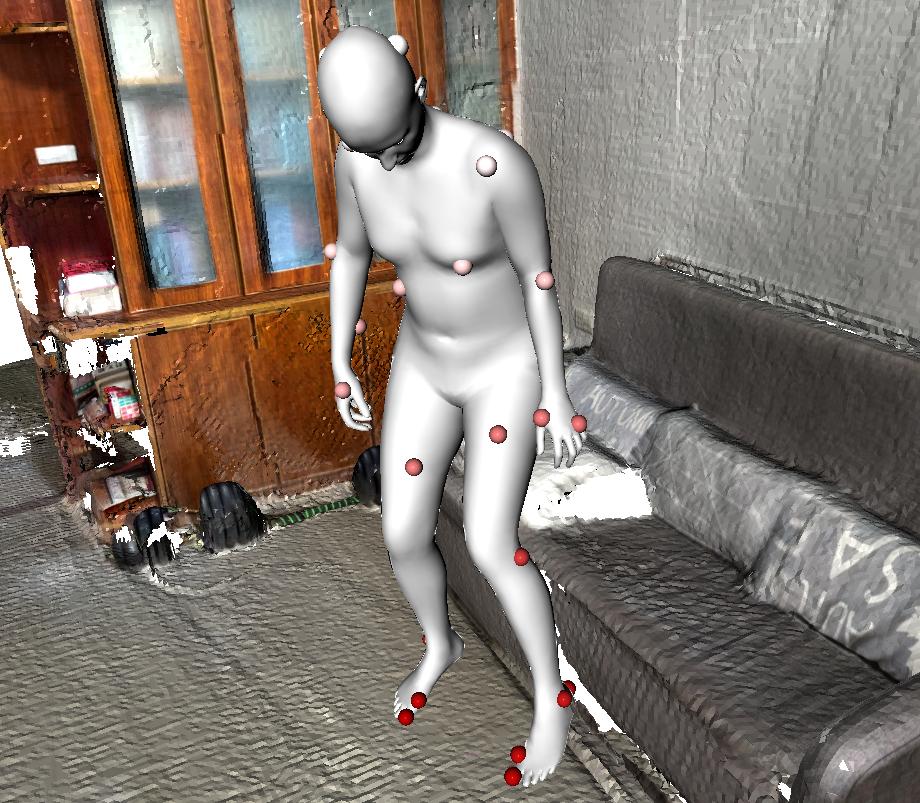} & \includegraphics[width=0.31\linewidth]{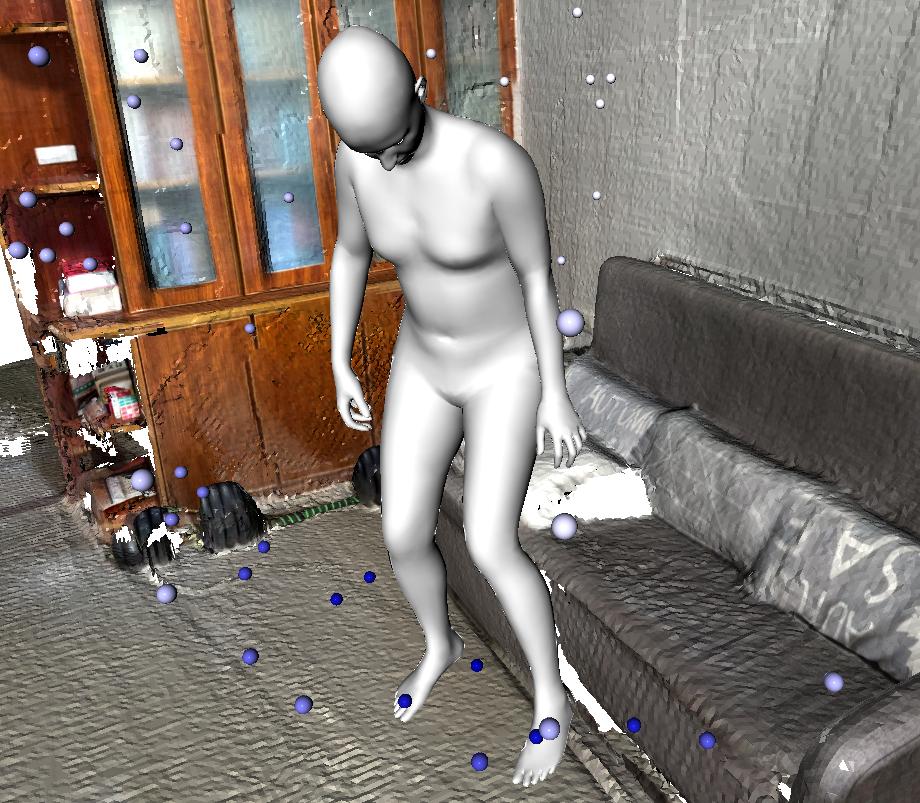}\\
          (a) The per-vertex signed distance & (b) The per-basis point distance
    \end{tabular}
    \caption{\textbf{Our mutual distances.} a) shows the per-vertex signed distance for sampled vertices on human mesh. Their color indicates the distance value. b) shows the per-basis point distance for the basis points which are not sampled on the scene surface. For both figures, the darker the color is, the smaller the distance is.}
    \label{fig:distances}
\end{figure*}
To constrain the whole-body motion for better prediction, in this paper, we propose a mutual distance representation which captures the distances between the human body and the scene. We further introduce the global scene representation learned from SDF volume instead of point cloud to achieve consistent global and local scene representation learning. Specifically, our proposed mutual distance representation comprises two components: the per-vertex signed distance and the per-basis point distance. Given a 3D scene, the per-vertex distance is the minimum distance from each human body vertex to the scene surface, and its sign indicates whether such vertex penetrates the scene or not (see Fig.~\ref{fig:distances} (a)). Since the change in signed distance for each body vertex represents the motion of the corresponding body part, it provides a complete constraint on the entire body.
However, as a signed distance corresponds to a level-set of the scene Signed Distance Field (SDF), a human mesh at multiple possible global locations can result in the same per-vertex signed distance.


To address this, we leverage a per-basis point distance, which calculates the minimum distance between each point in a predefined set of basis points and the human surface.
Our per-basis point distance is similar to BPS body encoding in PLACE~\cite{zhang2020place} with differences in choices of basis points. In particular, their basis points are partially scene-dependent making it not suitable for motion prediction, while our basis points are purely scene-independent and sampled with a fixed distance relative to origin, which can better capture the motion patterns. 
While our model is end-to-end trainable, we adopt a two-step training process followed by fine-tuning for better convergence.
In the first step, we predict the mutual distance.
Conditioning on the predicted distances,
we then forecast the future human motion.
Specifically, given a 3D scene, the past human poses, and past mutual distance, we use a strategy based on Graph Convolutional Network (GCN)~\cite{kipf2016semi}, Discrete Cosine Transform (DCT), for the spatial and temporal encoding of the mutual distances and a simple 3D CNN for encoding of the SDF volume to predict the future ones. We subsequently take the predicted distances, the past human poses, and the scene as input to predict future human motion using an RNN-based model. A consistency prior is adopted to ensure that the predicted human motion is consistent with the input mutual distance. 

Our contributions can be summarized as follows. i)We propose an explicit representation of human-scene interaction using mutual distance to provide constraints for whole-body motion. ii) We explore different global scene representations and find the SDF volume to be more efficient than commonly used point cloud representation. Our experiments demonstrate that our approach outperforms the state-of-the-art scene-aware human motion prediction methods by a large margin on both existing real and synthesis benchmark datasets.

%% file: 03-relatedwork.tex
\section{Related Work}\label{sec:related-work}
\textbf{Human motion prediction.} Researchers have been studying 3D human motion modeling~\cite{brand2000style,sidenbladh2002implicit,taylor2007modeling,wang2008gaussian} for decades. While traditional methods, such as those employing Hidden Markov Models~\cite{brand2000style} or the Gaussian process latent variable model~\cite{wang2008gaussian}, are capable of handling periodic and simple non-periodic motions like walking and golf swings, more complex motions are typically modeled with deep neural networks~\cite{Li_2018_CVPR,aksan2019structured,mao2019learning,wang2019imitation,gopalakrishnan2019neural,li2020dynamic,mao2020history,cai2020learning}. However, these methods focus solely on the prediction of local poses, neglecting the global human motion and 3D scenes in which the motion takes place.

More recent works~\cite{cao2020long, corona2020context, mao2022contact, zheng2022gimo, scofano2023staged} have begun to incorporate scene context into the task of motion prediction. Specifically, Cao~\etal~\cite{cao2020long} propose to represent the scene with a 2D image. The feature extracted from such an image is fed into the prediction model and lets the model learn the human-scene interaction implicitly. However, such 2D scene representation cannot handle occlusions and it does not ensure a valid human-scene interaction. In~\cite{corona2020context}, a semantic graph is introduced to encode the interaction between the human and the object of interest. Although it is efficient to model a human interacting with one object, it is not suitable for human-scene interaction as the human may interact with multiple objects when he moves around. In~\cite{zheng2022gimo},  Zheng~\etal employ eye gaze as additional information to guide the motion forecasting process. Although eye gaze provides valuable hints about the goal location of the human motion, it constrains neither the local poses nor the human-scene interaction. Mao~\etal~\cite{mao2022contact} introduce a contact map that can explicitly encode the relation between scene surface points and human joints. 
Scofano~\etal~\cite{scofano2023staged} propose a three-stage pipeline in a coarse-to-fine manner and utilizing the time information based on \cite{mao2022contact}.  
However, for forecasting, their contact map only constrains joints that contact the scene and disregard the rest. By contrast, our introduced mutual distance is able to constrain the whole-body motion.

\noindent\textbf{Scene-aware human synthesis.} One of the main challenges in this task is to ensure a valid human-scene interaction. As such, existing works have explored various strategies to model such interaction~\cite{zhang2020place,zhang2020generating,hassan2021populating,wang2021scene,wang2021synthesizing, hassan2021stochastic}. Similar to~\cite{cao2020long}, the interaction is modeled implicitly in~\cite{zhang2020generating,wang2021scene,wang2021synthesizing} via either Generative Adversarial Networks (GANs)~\cite{wang2021synthesizing}, or Variational Autoencoders~\cite{zhang2020generating,wang2021synthesizing}. Besides this, other works~\cite{zhang2020place,hassan2021populating,hassan2021stochastic} adopt explicit representation of human-scene interaction. In particular, it was proposed in POSA~\cite{hassan2021populating} to develop a human-centric representation of the human-scene interaction which assigns scene semantic labels, such as floor, and chair, to each human surface vertex. Those labels indicate the probability of such vertex contacting what kind of object in the scene. Although it performs well in the task of placing a human in a scene, it lacks sufficient detail about the local pose of the human, rendering it unsuitable for human motion forecasting. In SAMP~\cite{hassan2021stochastic}, Hassan~\etal propose to only model the interaction with the object at the goal location. Given the final root translation and orientation, the task is then treated as a path planning problem. Similar to~\cite{corona2020context}, such a strategy cannot constrain the interactions between the human and the scene when the human moves in the scene. The closest to our work is PLACE~\cite{zhang2020place} where the human-scene interaction is represented by the minimum distance between a set of 3D scene points also known as Basis Point Sets (BSP)~\cite{prokudin2019efficient} to the human body surface. However, 
such representation alone only provides a prior on the global position of the human. For example, one can move its arm that is far from the scene surface while the scene to human proximity remains the same.
By contrast, we combine the per-basis point distance with the per-vertex signed distance resulting in 
a detailed constraint on the whole-body motion both locally and globally. 

%% file: 04-methodology.tex
\section{Approach}
Let us begin by formulating the task of scene-aware 3D human motion forecasting. Building on previous work~\cite{mao2022contact}, we represent a sequence of $T$ past human poses as $\mathbf{X}=[\mathbf{x}_1,\mathbf{x}_2,\cdots,\mathbf{x}_T]\in \mathbb{R}^{M \times T}$\footnote{Unless otherwise specified, in this paper, we use column vector convention.}, where a human pose at time $t$ is represented as a vector $\mathbf{x}_t\in\mathbb{R}^{M}$ of length $M$. Such vector could be a parametric representation of human mesh such as SMPL-X~\cite{SMPL-X:2019} pose. 
To represent the 3D scene, we adopt a Signed Distance Volume $\mathbf{S}\in \mathbb{R}^{N \times N \times N}$, each entry of which stores the distance of the voxel to the nearest scene surface and its sign indicates whether it is inside or outside the scene. The representation learned from this volume is more consistent with our mutual distance representation than those learned from point cloud~\cite{mao2022contact}. Moreover, such Signed Distance Volume allows us to query the signed distance of any given position easily via interpolation, which facilitates the fast calculation of the loss defined on the SDFs of the predicted pose sequence during training (see Eq.~\ref{eq:per-vertex loss}).

\begin{figure*}[!t]
    \centering
    \includegraphics[width=\linewidth]{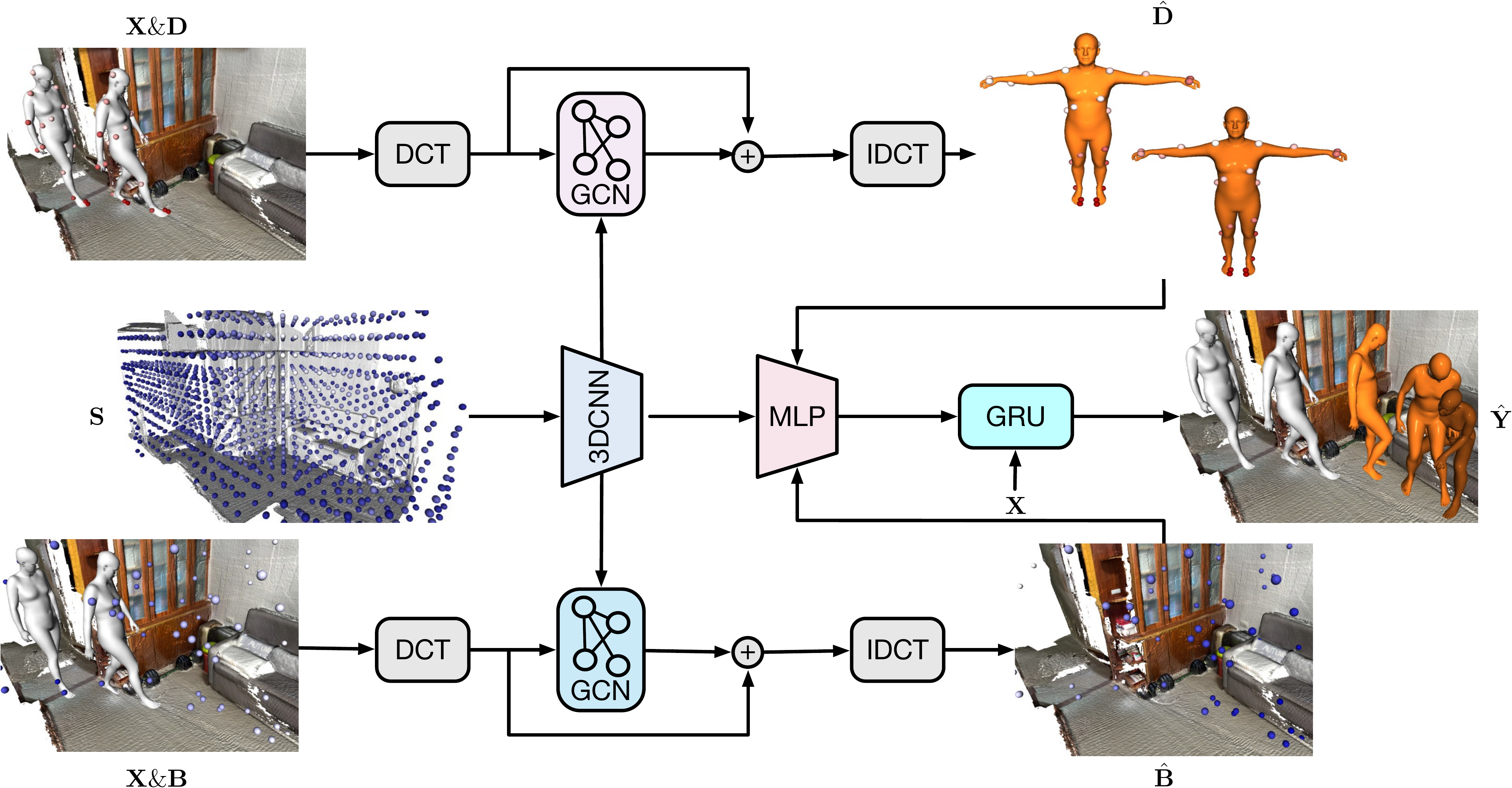}
    \caption{\textbf{Network architecture.} Given the 3D scene $\mathbf{S}$ represented as a signed distance volume and the past motion $\mathbf{X}$ shown in grey meshes, our approach first predicts the future per-vertex signed distance $\hat{\mathbf{D}}$, and the future per-basis point distance $\hat{\mathbf{B}}$ from historical distance $\mathbf{D}$ and $\mathbf{B}$, respectively. The two predicted future distances are then fed into a RNN-based network to predict the future motion $\hat{\mathbf{Y}}$ shown in orange meshes.}
    \label{fig:pipeline}
\end{figure*}

Our objective is predicting $U$ future human poses $\mathbf{Y}=[\mathbf{x}_{T+1},\mathbf{x}_{T+2},\cdots,\mathbf{x}_{T+U}]\in \mathbb{R}^{M \times U}$ with history motion $\mathbf{X}$ and 3D scene $\mathbf{S}$. We introduce an innovative pipeline that predicts mutual distance representation first i.e., the signed minimum distance from each human body vertex to the scene surface (the per-vertex signed distance) and the minimum distance from each basis point to the human surface (the per-basis point distance) at each time frame. Conditioned on such mutual distance, our model then forecasts future poses. Our pipeline is illustrated in Fig.~\ref{fig:pipeline}. We detail our mutual distance and the pipeline for predicting these distances and future motions below.

\subsection{Mutual Distance Representation}\label{sec:muDis}
Let us first introduce our mutual distance representation, which consists of the per-vertex signed distance and the per-basis point distance.

\textbf{The per-vertex signed distance.} Given the human pose $\mathbf{x}_t$ at time $t$, one can obtain the vertices of human mesh with a predefined mapping, such as SMPL-X~\cite{SMPL-X:2019}. 
However, constraining every single vertex may not be necessary due to their strong correlation. Hence, following~\cite{zhang2021we}, we select a subset of $K$ vertices denoted as $\mathbf{v}_t\in\mathbb{R}^{K\times 3}$ from the human mesh. Given a 3D scene and $\mathbf{v}_t$ at time $t$, the per-vertex signed distance can be computed as,
\begin{align}
        d_{kt} = \begin{cases}
        -\min\limits_{\mathbf{y}\in\partial\mathcal{S}} \|\mathbf{v}_{tk}-\mathbf{y}\|_2 \;&\text{if}\  \mathbf{v}_{tk} \in \mathcal{S}\\
        \;\;\;\min\limits_{\mathbf{y}\in\partial\mathcal{S}} \|\mathbf{v}_{tk}-\mathbf{y}\|_2 \;&\text{if}\  \mathbf{v}_{tk} \notin \mathcal{S} \;,
    \end{cases}
\end{align}
where $\mathbf{d}_t=[d_{1t},d_{2t},\cdots,d_{Kt}]^T\in \mathbb{R}^{K}$ is the per-vertex signed distance at time step $t$, $\mathcal{S}$ represents the volume inside the actual solid scene and $\partial\mathcal{S}$ is the surface points of such scene. The per-vertex signed distance provides information about the whole-body pose. However, the same human mesh at different global positions may result in the same per-vertex signed distance. To resolve such ambiguity, we further leverage the per-basis point distance which is described below.

\textbf{The per-basis point distance.} Given a set of basis points from the scene denoted as $\mathbf{p}\in\mathbb{R}^{P\times3}$, the per-basis point distance is simply the minimum distance in each basis point to the human mesh. 
As the human moves within the scene, its motion is reflected in the continuous change of
the per-basis point distance. Specifically, at time step $t$, it can be computed as,
\begin{equation}
    b_{pt} = \min_{\mathbf{y}\in\partial\mathcal{H}_t} \|\mathbf{p}_p-\mathbf{y}\|_2\;,\label{eq:pbpd}
\end{equation}
where $\partial\mathcal{H}_t$ is the human surface at time step $t$ and $\mathbf{b}_{t}=[b_{1t},b_{2t},\cdots,b_{Pt}]^T\in\mathbb{R}^{P}$. Note that our basis points are not necessary on the scene surface. 
In practice, for different motion sequences, our basis points can be fixed relatively to the root joint of the last observed pose. Such root joint is defined as the origin of that data sample. In particular, we employ the spherical Fibonacci lattice algorithm ~\cite{gonzalez2010measurement} to sample $P$ near-equidistant points from a sphere with radius $R$.
The per-basis point distance primarily captures the global location of the human, while the per-vertex signed distance constrains both the detailed whole-body pose and global location. By combining two distances, our mutual distance representation provides explicit constraints on both the whole-body local poses and the global human motion of the human.

Given the 3D scene $\mathbf{S}$, the past motion $\mathbf{X}$, we first pre-compute the past mutual distance denoted as $\mathbf{D}=[\mathbf{d}_1,\mathbf{d}_2,\cdots,\mathbf{d}_T]\in\mathbb{R}^{K \times T}$ and $\mathbf{B}=[\mathbf{b}_1,\mathbf{b}_2,\cdots,\mathbf{b}_T]\in\mathbb{R}^{P \times T}$. Our goal now is to predict the future mutual distance given the above information. In the following section, we will introduce our mutual distance prediction module. 


\subsection{Mutual Distance Prediction}
We adopt a similar temporal and spatial encoding strategy proposed in~\cite{mao2019learning}, which has been proven to be effective in predicting smooth spatial-temporal data such as 3D human motion. Specifically, we use Discrete Cosine Transform (DCT) to represent a temporal sequence and Graph Convolutional Networks (GCN) to capture the spatial dependencies between different sequences.

With the DCT, we represent a sequence as a linear combination of pre-defined cosine bases. In particular, let us denote a $T$-frame sequence of signed distance of the $k$-th human vertex as $\tilde{\mathbf{d}}_{k}\in[d_{1k},d_{2k},\cdots,d_{Tk}]^T\in\mathbb{R}^{T}$, the $l$-th DCT coefficient is then,
\begin{align}
    h_{kl} = \sqrt{\frac{2}{T}}\sum_{t=1}^{T}d_{tk}\frac{1}{\sqrt{1+\delta_{l1}}}\cos\left(\frac{\pi}{2T}(2t-1)(l-1)\right)\;,
    \label{eq:dct}
\end{align}
where $l\in \{1,2,\cdots T\}$ and $\delta_{ij}$ denotes the~\emph{Kronecker} delta function, i.e.,
\begin{equation}
  \delta_{ij} = \begin{cases}
  1 & \text{if}\ i=j\\
  0 & \text{if}\ i\neq j\;.
  \end{cases}
\end{equation}
Note that, with all cosine bases precomputed, the DCT can be simply represented as a linear operation,
\begin{equation}
    \mathbf{h}_{k} = \mathbf{C}\tilde{\mathbf{d}}_{k}\;,
\end{equation}
where $\mathbf{C}\in\mathbb{R}^{T\times T}$ is an orthogonal matrix and each row of $\mathbf{C}$ stores the precomputed the cosine basis. $\mathbf{h}_{k}\in\mathbb{R}^{T}$ are the $T$ DCT coefficients. Given such coefficients $\mathbf{h}_{k}$, the original sequence can be losslessly recovered via Inverse DCT (IDCT) as,
\begin{equation}   
    \tilde{\mathbf{d}}_{k} = \mathbf{C}^T\mathbf{h}_{k}\;.
\end{equation}

Similarly, we can obtain the DCT coefficients of all $K$ human vertices as $\mathbf{H}=[\mathbf{h}_1,\mathbf{h}_2,\cdots,\mathbf{h}_K]^T\in \mathbb{R}^{K\times T}$. For the spatial encoding, we follow~\cite{mao2019learning} to use GCN. Specifically, a GCN layer is defined as,
\begin{equation}
    \mathbf{F}^{(n+1)} = \sigma(\mathbf{A}^{(n)}\mathbf{F}^{(n)}\mathbf{W}^{(n)})\;,
\end{equation}
where $\sigma$ is an activation function such as $tanh(\cdot)$, $\mathbf{A}^{(n)}\in\mathbb{R}^{K\times K}$ is the learnable adjacent matrix at the $n$-th layer and $\mathbf{W}^{(n)}\in\mathbb{R}^{T\times F}$ is the feature extraction parameter. $\mathbf{F}^{(n)}\in\mathbb{R}^{K\times F'}$ and $\mathbf{F}^{(n+1)}\in\mathbb{R}^{K\times F}$ are the input and the output features of the $n$-th GCN layer, respectively\footnote{The first GCN layer will take the DCT coefficient matrix as input thus, $\mathbf{F}^{(1)} = \mathbf{H}$.}. Same as~\cite{mao2019learning}, our prediction network consists of many such layers. Since our final goal is to predict the future per-vertex signed distance, we follow the strategy in~\cite{mao2019learning} to first pad the sequence to make a sequence of length $T+U$. Our prediction module then aims to predict a residual between the ground truth DCT coefficients of the whole sequence and those of the padded historical sequence given the scene $\mathbf{S}$ and past human motion $\mathbf{X}$. More formally, the future per-vertex signed distance is predicted as follows,
\begin{equation}
    \hat{\mathbf{D}} = (\mathcal{G}_d(\hat{\mathbf{H}},\mathcal{G}_s(\mathbf{S}),\mathcal{G}_x(\mathbf{X})) + \hat{\mathbf{H}})\mathbf{C}\;,
\end{equation}
where $\mathcal{G}_d$, $\mathcal{G}_s$ and $\mathcal{G}_x$ are the GCN predictor, the 3D scene encoding network, and the past motion encoding model, respectively. With a little abuse of notation, here, $\mathbf{C}\in\mathbb{R}^{(T+U)\times(T+U)}$ is the DCT bases for sequences with length $T+U$. $\hat{\mathbf{H}}\in\mathbb{R}^{K\times (T+U)}$ is the DCT coefficients of the padded historical sequence.  $\hat{\mathbf{D}}\in\mathbb{R}^{K\times (T+U)}$ is the predicted per-vertex signed distance. Note that, here our model also tries to recover the past sequence. 

Similarly, we have another branch to predict the future per-basis point distance as,
\begin{equation}
    \hat{\mathbf{B}} = (\mathcal{G}_b(\hat{\mathbf{K}},\mathcal{G}_s(\mathbf{S}),\mathcal{G}_x(\mathbf{X})) + \hat{\mathbf{K}})\mathbf{C}\;,
\end{equation}

where $\mathcal{G}_b$ is also a GCN. $\hat{\mathbf{K}}$ is the DCT coefficients of padded past per-basis point distance sequence. $\hat{\mathbf{B}}\in\mathbb{R}^{P\times (T+U)}$ is the predicted per-basis point distance.

During training, we use the average $\ell_1$ loss between the ground-truth distances and the predicted ones. Formally, this loss is defined as,
\begin{equation}
    \ell_{\text{dist}}= \frac{1}{(T+U)}(\frac{1}{K}\sum_{t=1}^{T+U}\sum_{k=1}^{K}|
    \hat{d}_{kt}-d_{kt}|+ \frac{1}{P}\sum_{t=1}^{T+U}\sum_{p=1}^{P}|
    \hat{b}_{pt}-b_{pt}|)\;,\label{eq:loss_dist}
\end{equation}
where $\hat{d}_{kt}$ is predicted signed distance for $k$-th vertex at time $t$ and $\hat{b}_{pt}$ is predicted distance for $p$-th basis point at time $t$, $d_{kt}$ and $b_{pt}$ define the ground truth.

\subsection{Motion Forecasting Network}
Given the future mutual distance $\hat{\mathbf{D}}$ and $\hat{\mathbf{B}}$, the past human motion $\mathbf{X}$, and the scene $\mathbf{S}$, our goal now is to predict the future human poses $\hat{\mathbf{Y}}$. To this end, we use an RNN-based motion prediction model ($\mathcal{G}_{y}$) which predicts the future motion in an auto-regressive manner. More formally, it is defined as,
\begin{align}
    \hat{\mathbf{x}}_{T+u} = \mathcal{G}_{y}(\hat{\mathbf{x}}_{T+u-1},\hat{\mathbf{d}}_{T+u},\hat{\mathbf{b}}_{T+u},\mathcal{G}_x(\mathbf{X}),\mathcal{G}_s(\mathbf{S}))\;,
\end{align}
where $u\in\{1,2,\cdots,U\}$, $\hat{\mathbf{x}}_{T+u}\in\mathbb{R}^{M}$ is the predicted $u$-th future human pose. 

During training, in addition to the human pose loss, we incorporate two extra losses to ensure the predicted future pose to be consistent with the \emph{ground truth} mutual distance. For human pose loss, we employ global translation and orientation loss and the local pose loss. They are defined as follows,
\begin{align}
    \ell_{\text{global}} &= \frac{1}{U}\sum_{u=1}^{U}|\hat{\mathbf{t}}_{T+u}-\mathbf{t}_{T+u}| + |\hat{\mathbf{r}}_{T+u}-\mathbf{r}_{T+u}|\\
    \ell_{\text{local}} &= \frac{1}{U}\sum_{u=1}^{U}|\hat{\mathbf{x}}^{\text{local}}_{T+u}-\mathbf{x}^{\text{local}}_{T+u}|\;,
\end{align}


where $\hat{\mathbf{t}}_{T+u}\in\mathbb{R}^3$ and $\mathbf{t}_{T+u}\in\mathbb{R}^3$ are the predicted and ground truth global translation, respectively. $\hat{\mathbf{r}}_{T+u}\in\mathbb{SO}(3)$ and $\mathbf{r}_{T+u}\in\mathbb{SO}(3)$ are the predicted and ground truth global orientation, respectively. Here, we adopt the 6-D rotation representation~\cite{zhou2019continuity}. $\hat{\mathbf{x}}^{\text{local}}_{T+u}\in\mathbb{R}^{M-9}$ and $\mathbf{x}^{\text{local}}_{T+u}\in\mathbb{R}^{M-9}$ are the predicted and ground truth local human pose. To compute the two consistency losses, we first use the pre-defined mapping function such as SMPL-X~\cite{SMPL-X:2019} to obtain the sampled human surface vertices $\hat{\mathbf{v}}_{T+u}\in\mathbb{R}^{K\times 3}$  and the set of all human surface points $\partial\hat{\mathcal{H}}_{T+u}$.
Recall that we represent the 3D scene as a signed distance volume. We can query the signed distance of the human surface vertices by interpolating within this volume. The per-vertex signed distance consistency loss is then defined as,
\begin{align}
    \ell_{\text{vertex}} = \frac{1}{UK}\sum_{u=1}^{U}\sum_{k=1}^{K}|\tilde{d}_{k(T+u)}-d_{k(T+u)}|\;,
    \label{eq:per-vertex loss}
\end{align}
where $\tilde{d}_{k(T+u)}$ and $d_{k(T+u)}$ are the signed distance of the $k$-th vertex from the $u$-th predicted human mesh and ground truth human mesh, respectively.

Similarly, given the set of human surface points $\partial\hat{\mathcal{H}}_{T+u}$ at time $T+u$, we can compute the per-basis point distance $\tilde{\mathbf{b}}_{T+u}\in\mathbb{R}^{P}$ using Equation~\ref{eq:pbpd}. The per-basis point distance consistency loss can then be computed as
\begin{align}
    \ell_{\text{basis}} = \frac{1}{UP}\sum_{u=1}^{U}\sum_{p=1}^{P}|\tilde{b}_{p(T+u)}-b_{p(T+u)}|\;.
\end{align}

The final training loss for the motion forecasting network is the combination of all the above losses,
\begin{align}
    \ell_{\text{motion}} = \lambda_1 \ell_{\text{global}} +  \lambda_2 \ell_{\text{local}} + \lambda_3 \ell_{\text{vertex}} + \lambda_4\ell_{\text{basis}}\;,\label{eq:loss_motion}
\end{align}
where $\lambda_{i}$ with $i\in\{1,2,3,4\}$ are the balancing weights.

Here we adopt a stage-wise training strategy where we initially train the mutual distance prediction network and motion forecasting network separately. 
While our network is differentiable and can be trained from scratch in an end-to-end fashion, we observed that at early stage of the training process, the learning process can easily converge to a local minimum due to inaccurate prediction of mutual distance which serves as the input for the motion forecasting network. We thus propose to perform stage-wise training first.
Note that, for stage-wise training process, the motion forecasting network is provided with the ground truth mutual distance as input. Subsequently, we combine both networks and fine-tune them in an end-to-end manner with both losses defined in Eq.~\ref{eq:loss_dist} and~\ref{eq:loss_motion}.

%% file: 05-expriments.tex
\section{Experiments}\label{sec:exp}

\subsection{Datasets}
We evaluate our method on four datasets, HUMANISE~\cite{wang2022humanise}, GTA-IM~\cite{cao2020long}, PROX~\cite{hassan2019resolving} and GIMO~\cite{zheng2022gimo}. 
\noindent\textbf{GTA-IM.} The GTA Indoor Motion dataset~\cite{cao2020long}, building upon GTA game engine, is a synthetic human-scene interaction dataset. It comprises approximately 1 million RGB-D frames featuring 3D human poses for 50 distinct characters across 7 diverse scenes. To reconstruct the mesh, we fuse the depth maps of each scene and subsequently generate the SDF volume. We follow ContactAware~\cite{mao2022contact} to use the same training-test split and 21 joints. The videos are 30 FPS. 
Both our model and baseline models are trained to observe the past 30 frames to predict the future 60 frames following~\cite{mao2022contact}.

\noindent\textbf{PROX.} Proximal Relationships with Object eXclusion (PROX)~\cite{hassan2019resolving} is a real-world dataset. This dataset contains human SMPL-X parameters~\cite{SMPL-X:2019} fitted from RGB-D frames and 20 distinct characters interacting with various scenes. Since the original motion sequences are jittery, we use the refined version provided by~\cite{mao2022contact}. We adopt the same data split and the 22 joints as in~\cite{mao2022contact}. The frame rate is 30 Hz. 
Our model and baseline models input the past 30 frames to forecast the future 60 frames following~\cite{mao2022contact}.

\noindent\noindent\textbf{HUMANISE.} The HUMANISE dataset~\cite{wang2022humanise} is a synthetic human-scene interaction dataset. It is mainly built upon two primary datasets: the AMASS dataset~\cite{AMASS:2019} and the ScanNet dataset~\cite{dai2017scannet}. HUMANISE combines these datasets by incorporating different global motions to meet contact and collision constraints. Since the original split includes duplicated human motions in the training and test set which is not suitable for motion prediction, we create a new split to ensure that motions in the test set are never observed during training.~Moreover, we ensure that the scenes paired with the motions in the test set are categorized into~\emph{seen scenes}, which include a set of scenes observed during the training, and~\emph{unseen scenes} consisting of scenes never observed during training. Our final split comprises approximately 20k sub-sequences for training and 6k for testing, which includes 5k sub-sequences from seen scenes and 1k from unseen scenes.
The videos run at 30 frame-per-second (FPS). 
Both our model and baseline models are trained to observe past 15 frames (0.5 sec.) and predict future 30 frames (1 sec.) because the average length of motion sequences in the dataset is 2.2 sec. which is much shorter than 3 sec.

\noindent\textbf{GIMO.} The Gaze-Informed Human Motion dataset (GIMO)~\cite{zheng2022gimo} is a real human-scene interaction dataset. The dataset captures human motion using an IMU-based motion capture system, and the 3D scene is scanned using a smartphone equipped with lidar sensors. Same as~\cite{wang2022humanise}, GIMO also adopts SMPL-X~\cite{SMPL-X:2019} representation. It captures 11 subjects performing various actions e.g., open window, lie in bed, in 19 scenes, resulting in 127 long motion sequences with 192k frames. We follow the official training and test split, resulting in around 6k training sequences and 1k testing sequences. Same as HUMANISE~\cite{wang2022humanise}, frame rate is 30Hz. Our model and baseline models predict future 30 frames with 15 past frames.



\subsection{Metrics, Baselines \& Implementation}
\noindent\textbf{Metrics.} We follow~\cite{mao2022contact} to evaluate our method using path error and pose error. The path error is defined as the $l_2$ distance between the predicted and the ground truth translation; the pose error is the Mean Per Joint Position Error (MPJPE)~\cite{h36m_pami} of local poses.

\noindent\textbf{Baselines.} We compare our method with three existing scene-aware human motion prediction models: ContAware~\cite{mao2022contact}, GIMO~\cite{zheng2022gimo} and STAG~\cite{scofano2023staged}. ContAware~\cite{mao2022contact} and STAG~\cite{scofano2023staged} are state-of-the-art methods which first predict the contact map and then the future motion conditioning on the predicted contact map. We train their model using the official implementation on our datasets, following their original settings to predict only human joints instead of SMPL-X poses. GIMO~\cite{zheng2022gimo} is another state-of-the-art scene-aware human motion prediction method which mainly focuses on predicting long-term future poses given the 
past gaze information, and the frame rate of their motion is 2Hz. The original model of GIMO predicts future 10 frames (5 seconds) given the past 6 frames (3 seconds). We slightly adapt their official code to our settings for evaluation.

\noindent\textbf{Implementation details.} Our models are developed using Pytorch~\cite{paszke2017automatic} and optimized with the ADAM~\cite{kingma2014adam} optimizer. Initially, we train the mutual distance prediction model and the motion forecasting model separately for 40 epochs each. Subsequently, we fine-tune both models together for an additional epoch. The original learning rate is set to 0.0005 and decays by half at the 30-th epoch for both models. We train our model on a single RTX4090. For all datasets, the loss weights of the motion forecasting model ($\lambda_1$,$\lambda_2$,$\lambda_3$,$\lambda_4$) are $1.0, 0.5, 1.0, 1.0$. For each motion sequence, we crop the 3D scene to a region that is within 2.0 meters of the root joint of the last observed pose, and the root joint is used as the origin of the cropped scene. To compute the per-vertex signed distance, we follow~\cite{zhang2021we} to use 67 vertices on the human body. For the basis point set, we select 150 points that are 2.0 meters away from the origin of the scene, and those points are fixed for different sequences.

\begin{table}[!ht]
    \centering    
    \resizebox{\linewidth}{!}{
    \begin{tabular}{cc ccccc c ccccc}
    \toprule
    & & \multicolumn{5}{c}{Path Error (mm)} & & \multicolumn{5}{c}{Pose Error (mm)}\\\cmidrule{3-7}\cmidrule{9-13}
    dataset & method & 0.5s & 1.0s & 1.5s & 2.0s & mean & &0.5s & 1.0s& 1.5s& 2.0s &  mean\\
    \midrule \parbox[t]{2mm}{\multirow{7}{*}{\rotatebox[origin=c]{90}{GTA-IM}}}
    & ContAware~\cite{mao2022contact}$^\dagger$& 44.5&	82.6&	125.6&	182.9&	87.1& & 40.1&	54.1&	65.2&	77.2 & 51.8\\
    & GIMO~\cite{zheng2022gimo}$^*$ & 52.7	& 97.8 &	160.6&	241.7&	110.3& & 47.9&	60.7&	71.1&	82.7&	59.9\\
    & STAG~\cite{scofano2023staged}$^\dagger$ &43.2&	79.8&119.9&176.4&83.4& &35.4&48.7&59.8&73.5&47.0\\
    & Ours & \textbf{34.4} & \textbf{65.9} & \textbf{104.0} & \textbf{155.6} & \textbf{72.0} &  & \textbf{31.0} & \textbf{46.8} & \textbf{58.9} & \textbf{70.7} & \textbf{44.6}\\
    \cmidrule{2-13}
    & ContAware w/GT &34.5&	46.0&	49.7&	67.6&	41.7& &	36.3&	44.3&	48.7&	53.9&	41.3\\
    & STAG w/GT &32.7&	43.3&	49.4&	68.1&	40.0& &31.0&	37.2&	41.5&	48.3&	35.1\\
    & Ours w/GT &\textbf{15.8} &\textbf{18.9}&	\textbf{20.5}&	\textbf{23.0}&	\textbf{22.4}& &\textbf{20.4}&	\textbf{23.8}&	\textbf{25.8}&	\textbf{28.2}&	\textbf{17.6}\\
    \midrule \parbox[t]{2mm}{\multirow{7}{*}{\rotatebox[origin=c]{90}{PROX}}}
    & ContAware~\cite{mao2022contact}& 93.3 & 187.2 & 284.4 & 381.2 & 192.2 & & 89.9 & 127.5 & 149.3 & 167.5 &116.8 \\
    & GIMO~\cite{zheng2022gimo}$^*$ & 129.5 & 233.2 & 342.1 &455.0 & 240.4 & & 112.6 &144.6 & 168.0 & 187.5 &140.7 \\
    & STAG~\cite{scofano2023staged}$^\dagger$ & 95.9 &	186.4&	276.6&	369.8&	189.6& &	98.4&	137.8&	163.7&	185.7&	129.4 \\
    & Ours & \textbf{80.9} & \textbf{155.2} & \textbf{238.6} & \textbf{322.1} & \textbf{162.6} &  & \textbf{83.9} & \textbf{119.3} & \textbf{142.0} & \textbf{159.7} & \textbf{110.7} \\
    \cmidrule{2-13}
    & ContAware w/GT & 73.9&	106.7&	104.6&	117.4& 	88.0& &	83.7&	112.9&	125.2&	132.9&	101.1\\
    & STAG w/GT& 64.0 &	81.6 &	93.4 &	115.4 &	77.2 & &	88.6 &	106.0 &	112.6 &	125.4&	98.3\\
    & Ours w/GT & \textbf{44.2}& 	\textbf{56.7}&	\textbf{63.7}&	\textbf{69.9}&	\textbf{52.6}& &	\textbf{56.4}&	\textbf{69.4}&	\textbf{76.8}&	\textbf{83.4}&	\textbf{64.7}\\
    \bottomrule
    \end{tabular}
}
    \caption{\textbf{Quantitative results on GTA-IM~\cite{cao2020long} and PROX~\cite{hassan2019resolving}.} Our model outperforms all baselines at all metrics, especially for the path error. }
    \label{tab:gtaim}
\end{table}

\begin{table}[!ht]
    \centering
    \resizebox{\linewidth}{!}{
        \begin{tabular}{c ccc c ccc c ccc c ccc}
        \toprule
        & \multicolumn{7}{c}{Seen Scenes} & & \multicolumn{7}{c}{Unseen Scenes}\\
        \cmidrule{2-8} \cmidrule{10-16}
        & \multicolumn{3}{c}{Path Error (mm)} & & \multicolumn{3}{c}{Pose Error (mm)} & & \multicolumn{3}{c}{Path Error (mm)} & & \multicolumn{3}{c}{Pose Error (mm)} \\
        \cmidrule{2-4} \cmidrule{6-8} \cmidrule{10-12} \cmidrule{14-16}
        Method & 0.5s & 1.0s & mean & & 0.5s & 1.0s & mean & & 0.5s & 1.0s & mean & & 0.5s & 1.0s & mean \\
        \midrule
        ContAware~\cite{mao2022contact}$^\dagger$ & 52.8 & 121.1 & 57.8 & & 97.6 & 141.4 & 92.9 & & 53.1 & 124.0 & 58.7 & & 94.0 & 139.1 & 90.3 \\
        GIMO~\cite{zheng2022gimo}$^*$ & 70.1 & 129.2 & 72.0 &  & 141.4 & 150.3 & 140.2& & 77.7 & 144.0 & 80.2 &  & 146.3 & 159.4 & 146.5  \\
        STAG~\cite{scofano2023staged}$^\dagger$ & 55.2&	124.6&	60.7& &		88.3&	131.4&	83.0& & 57.0&	131.5&	63.2& &		89.9&	137.7&	85.6\\
        Ours & \textbf{41.5} & \textbf{93.5} & \textbf{45.6} &  & \textbf{83.7} & \textbf{130.9} & \textbf{80.0} &  & \textbf{46.7} & \textbf{100.2} & \textbf{50.1} &  & \textbf{84.3} & \textbf{131.8} & \textbf{80.6} \\
        \midrule
        ContAware w/GT& 38.0 &	60.8&	35.7& &93.4&	119.1&	86.4& & 38.2&	61.8&36.4& & 90.3&	122.0&	84.2\\
        STAG w/GT& 37.5&	63.2&	36.4& &		83.5&	\textbf{117.6}&	77.3& & 37.7&	62.2&	36.6& &		85.1&	122.7&	79.4\\
        Ours w/GT& \textbf{22.9} &	\textbf{33.4}&	\textbf{22.0}& &\textbf{80.3}&	122.0&	\textbf{76.3}& &\textbf{24.8}&	\textbf{35.5}&	\textbf{23.6}& &		\textbf{80.2}&	\textbf{121.5}&	\textbf{76.2}\\
        \bottomrule
        \end{tabular}}
    \caption{\textbf{Quantitative results on HUMANISE~\cite{wang2022humanise}.} Our model outperforms all baselines on both seen scenes and unseen scenes.}
    \label{tab:humanise}
\end{table}

\begin{table}[!ht]
    \centering
    \resizebox{\linewidth}{!}{
        \begin{tabular}{c ccc c ccc c ccc c ccc}
        \toprule
        & \multicolumn{7}{c}{Seen Scenes} & & \multicolumn{7}{c}{Unseen Scenes}\\
        \cmidrule{2-8} \cmidrule{10-16}
        & \multicolumn{3}{c}{Path Error (mm)} & & \multicolumn{3}{c}{Pose Error (mm)} & & \multicolumn{3}{c}{Path Error (mm)} & & \multicolumn{3}{c}{Pose Error (mm)} \\
        \cmidrule{2-4} \cmidrule{6-8} \cmidrule{10-12} \cmidrule{14-16}
        Method & 0.5s & 1.0s & mean & & 0.5s & 1.0s & mean & & 0.5s & 1.0s & mean & & 0.5s & 1.0s & mean \\
        \midrule
        Motion only& 53.4 & 111.9 & 57.0 & & 102.0 & 140.0 & 93.5 & & 59.2 & 119.0 & 62.0 & & 103.2 & 145.3 & 95.5 \\
        Scene only& 48.4 & 105.8 & 52.6 & & 94.6 & 132.9 & 87.6 & & 54.3 & 113.8 & 57.9 & & 94.1 & 136.9 & 88.3\\
        Ours w/o $\mathbf{D}$ & 42.4 & 95.2& 46.2& & 88.2& 134.6& 83.4 & & 47.9 & 103.1& 51.2& & 90.4& 140.1& 86.0\\
        Ours w/o $\mathbf{B}$ & 47.6 & 101.3 & 50.6 & & 85.7 & 133.9 & 81.7& & 52.8 & 109.1 & 55.8 & & 86.4 & 138.3 & 83.2\\
        Ours w/ $P$ & 45.2&	97.2&	49.3& & 90.8&	137.2&	86.0& & 49.0&	101.7&	52.7& & 89.8&	138.9&	86.1\\
        Ours w/ $O$ & 50.1&115.3&55.3& &92.2&141.2&86.7& & 55.0 & 115.2 & 58.4& & 91.0 & 141.3 & 86.2\\
        Ours w/ $BPS$ & 50.2 & 108.9& 53.9& & 100.0 & 153.6 & 94.5 & & 52.0 & 104.1 & 54.3 & & 94.4 & 148.2 & 90.3\\
        Ours $e2e$ & 47.6&	105.2&	51.5 & &		90.4&	141.0&	86.2& &52.1&	112.4&	56.1& &		89.5&	142.0	&85.7\\
        Ours & \textbf{41.5} & \textbf{93.5} & \textbf{45.6} &  & \textbf{83.7} & \textbf{130.9} & \textbf{80.0} &  & \textbf{46.7} & \textbf{100.2} & \textbf{50.1} &  & \textbf{84.3} & \textbf{131.8} & \textbf{80.6} \\
        \bottomrule
        \end{tabular}}
    \caption{\textbf{Ablation study on HUMANISE~\cite{wang2022humanise}.} We show the ablation results of our model. $\mathbf{D}$, $\mathbf{B}$, $P$, $O$, $BPS$ and $e2e$ indicate per-vertex signed distance, per-basis point distance, point cloud representation, occupancy representation, BPS encoding and end-to-end training respectively. }
    \label{tab:ablation} 
\end{table}

\begin{figure}
\centering
\begin{tabular}{cccccc}
& Ground Truth & Ours & GIMO~\cite{zheng2022gimo} & ContAware~\cite{mao2022contact} & STAG\cite{scofano2023staged} \\ 
\multicolumn{1}{c}{\rotatebox{90}{\parbox[c]{2cm}{\centering GTA-IM}}} & \includegraphics[width=0.2\textwidth]{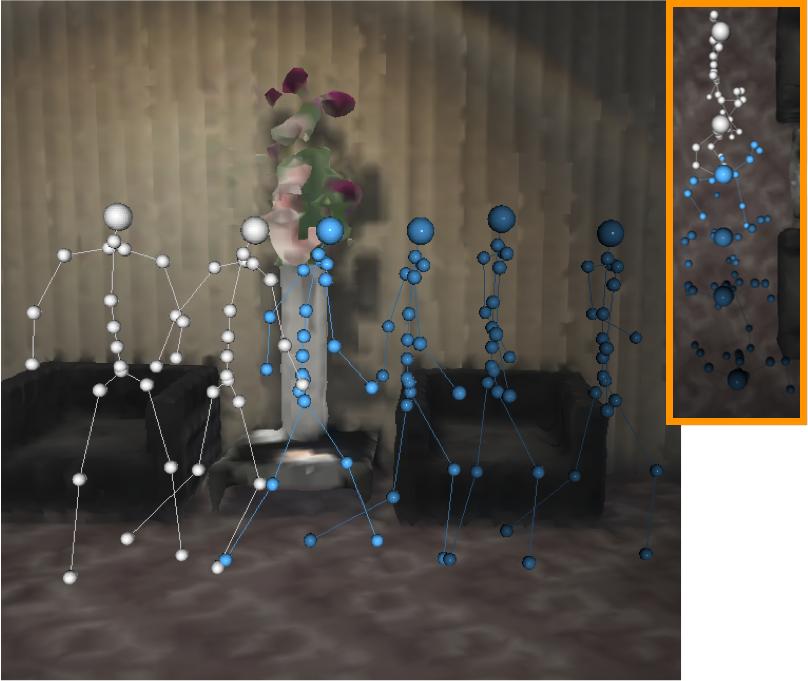}\hspace{-1mm} & \includegraphics[width=0.2\textwidth]{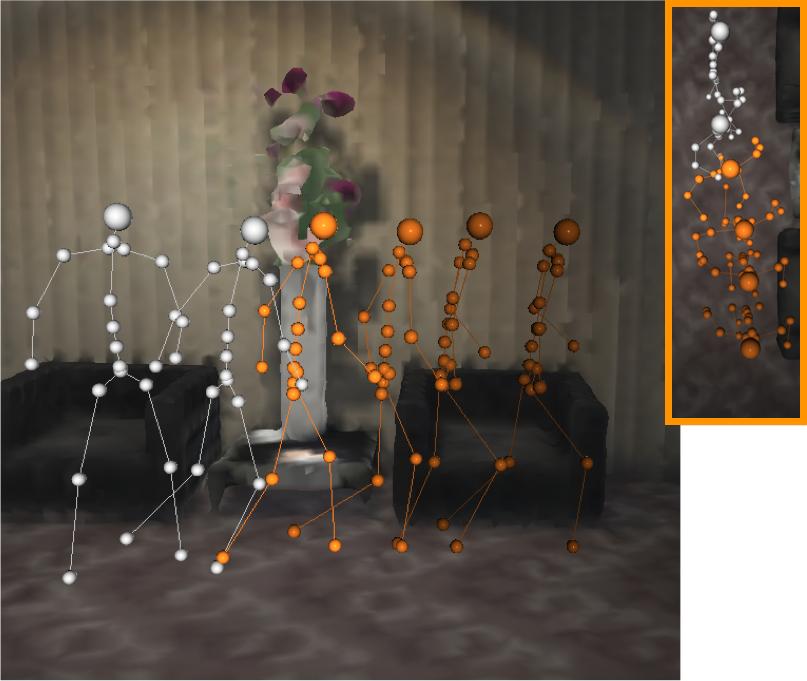}\hspace{-1mm} & \includegraphics[width=0.2\textwidth]{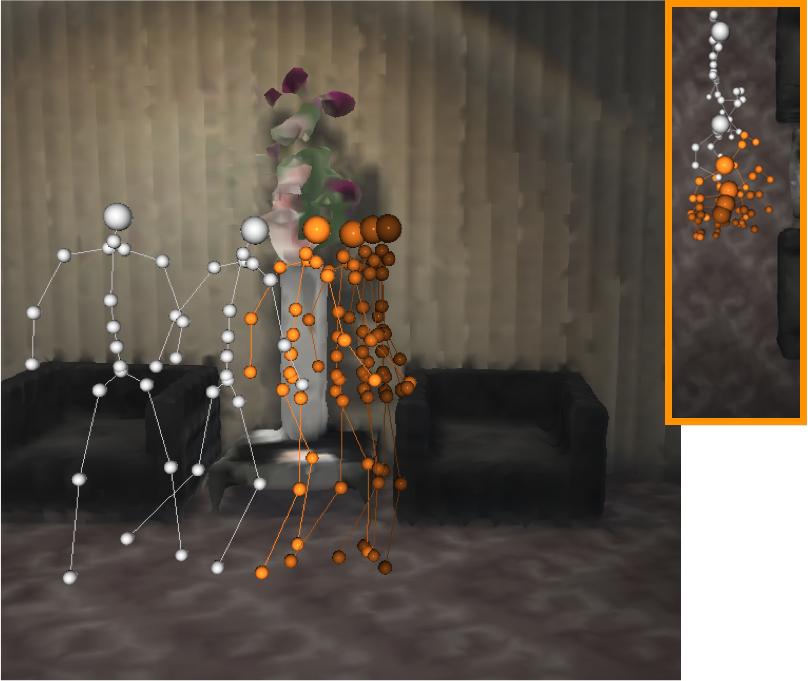}\hspace{-1mm} & \includegraphics[width=0.2\textwidth]{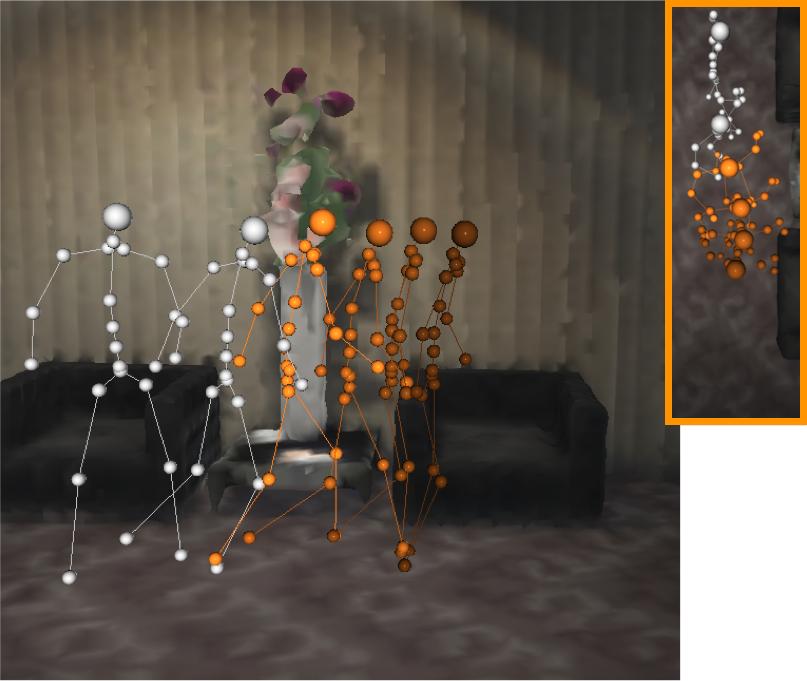}\hspace{-1mm} & \includegraphics[width=0.2\textwidth]{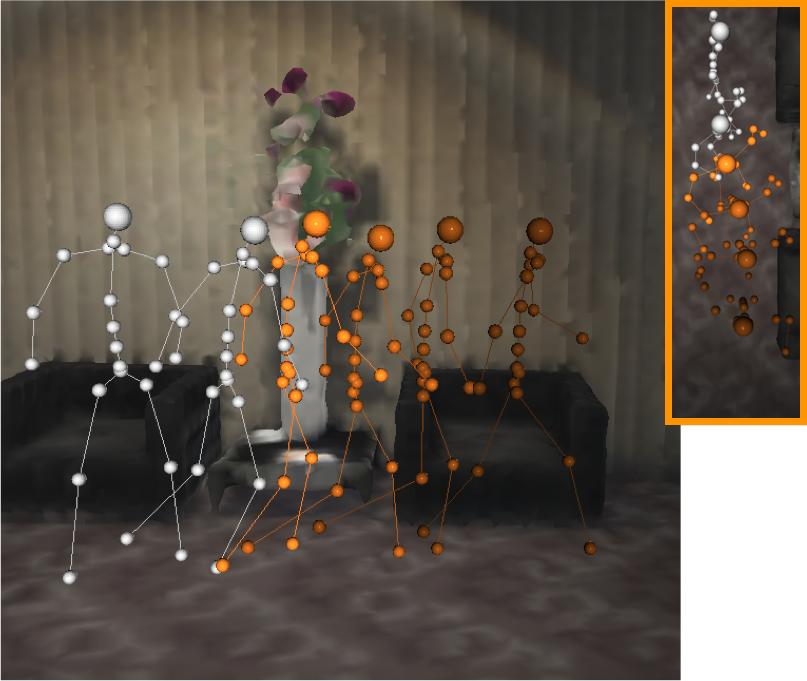}\hspace{-1mm} \\
\multicolumn{1}{c}{\rotatebox{90}{\parbox[c]{2cm}{\centering PROX}}} & \includegraphics[width=0.2\textwidth]{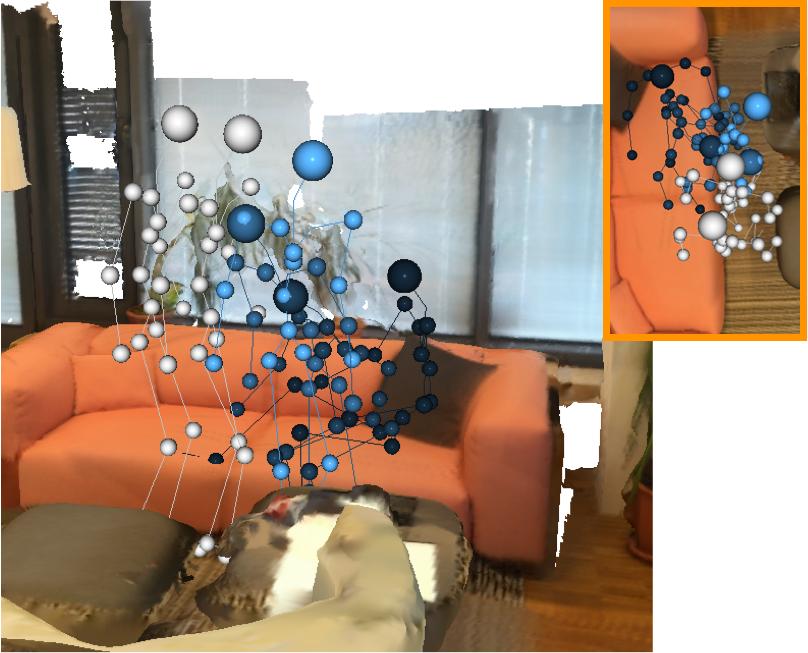}\hspace{-1mm}& \includegraphics[width=0.2\textwidth]{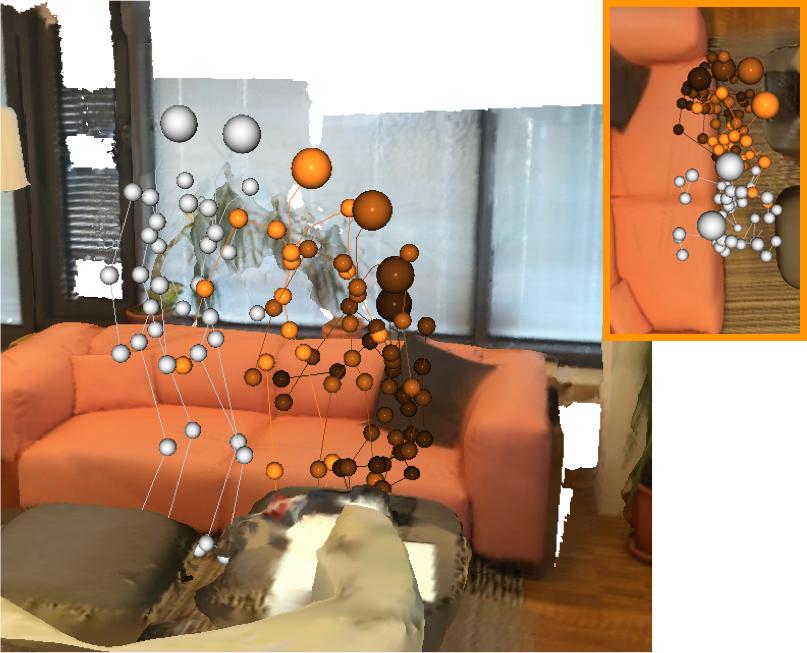}\hspace{-1mm} & \includegraphics[width=0.2\textwidth]{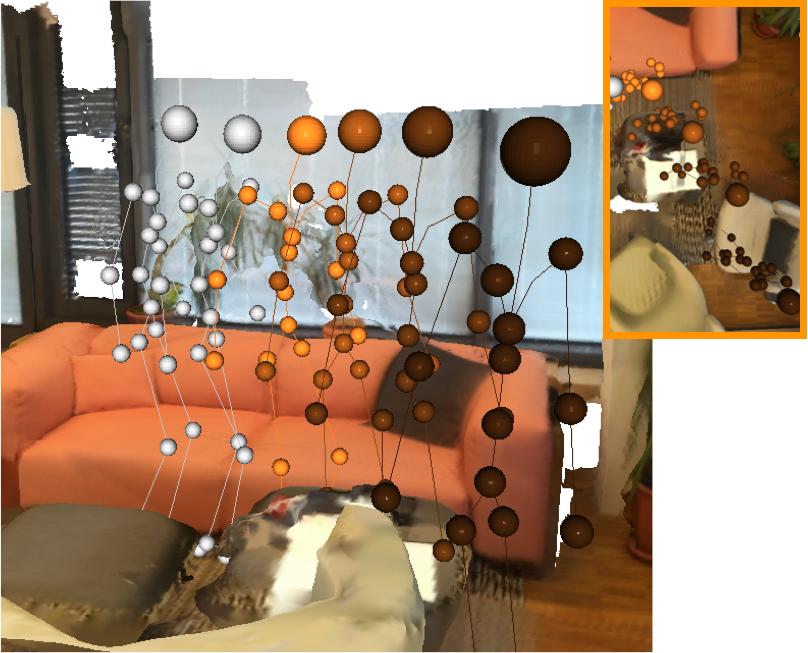}\hspace{-1mm} & \includegraphics[width=0.2\textwidth]{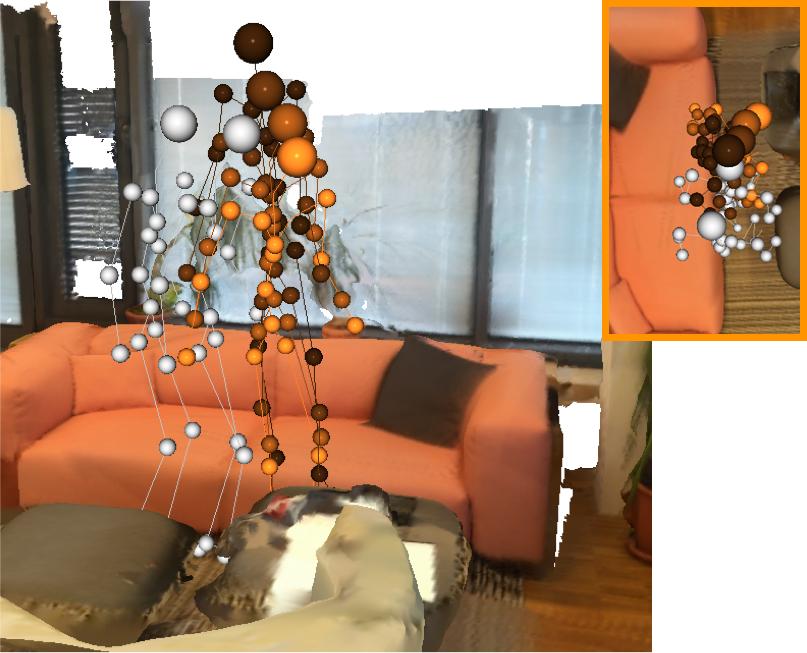}\hspace{-1mm} & \includegraphics[width=0.2\textwidth]{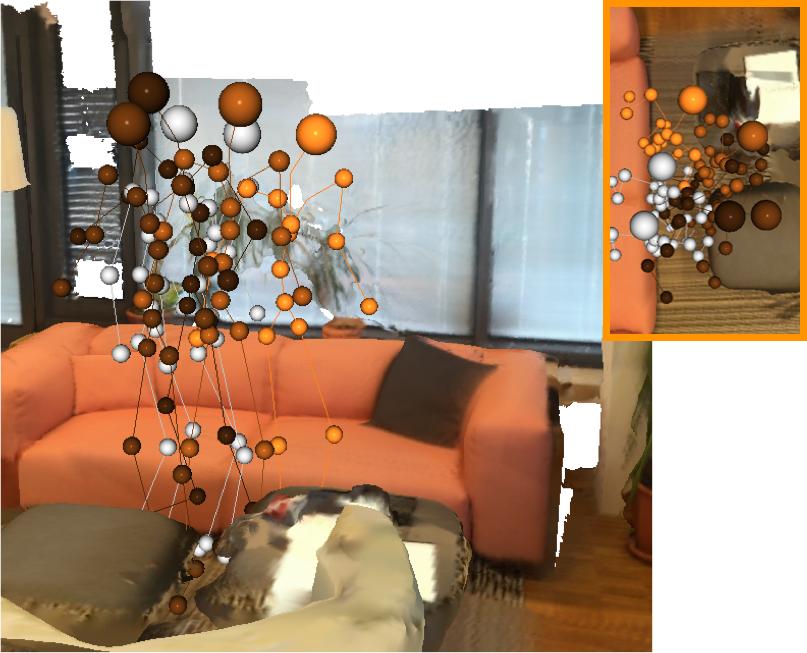}\hspace{-1mm} \\
\multicolumn{1}{c}{\rotatebox{90}{\parbox[c]{2cm}{\centering {HUMANISE}}}} & \includegraphics[width=0.2\textwidth]{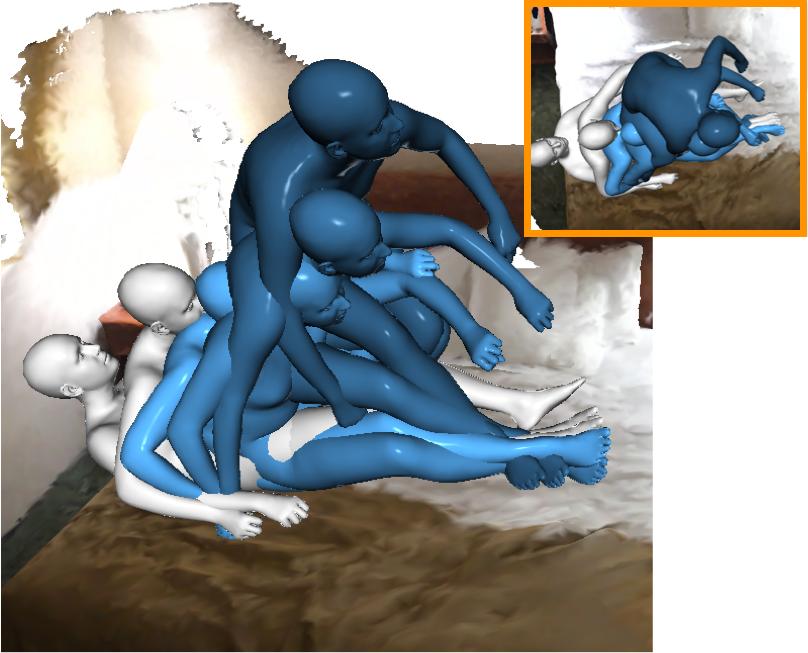}\hspace{-1mm} & \includegraphics[width=0.2\textwidth]{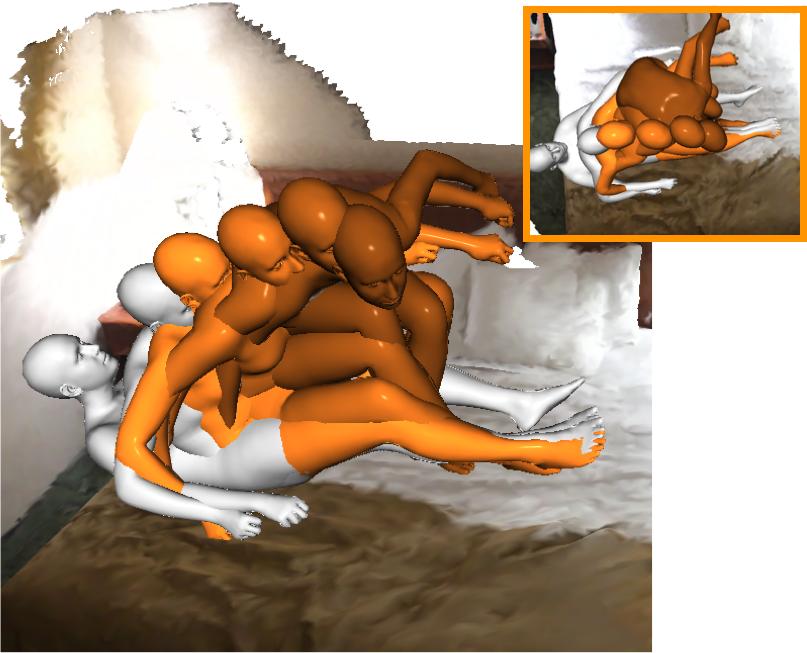}\hspace{-1mm} & \includegraphics[width=0.2\textwidth]{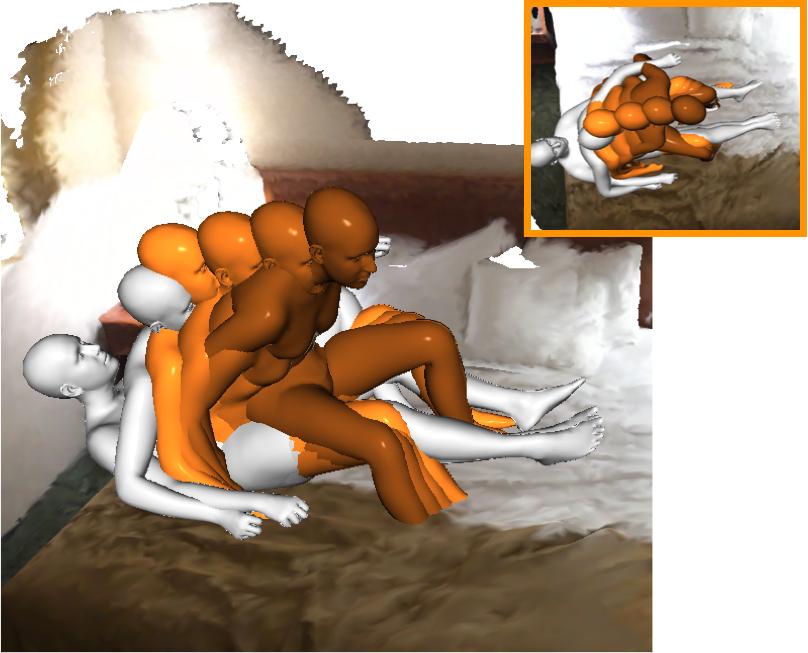}\hspace{-1mm} & \includegraphics[width=0.2\textwidth]{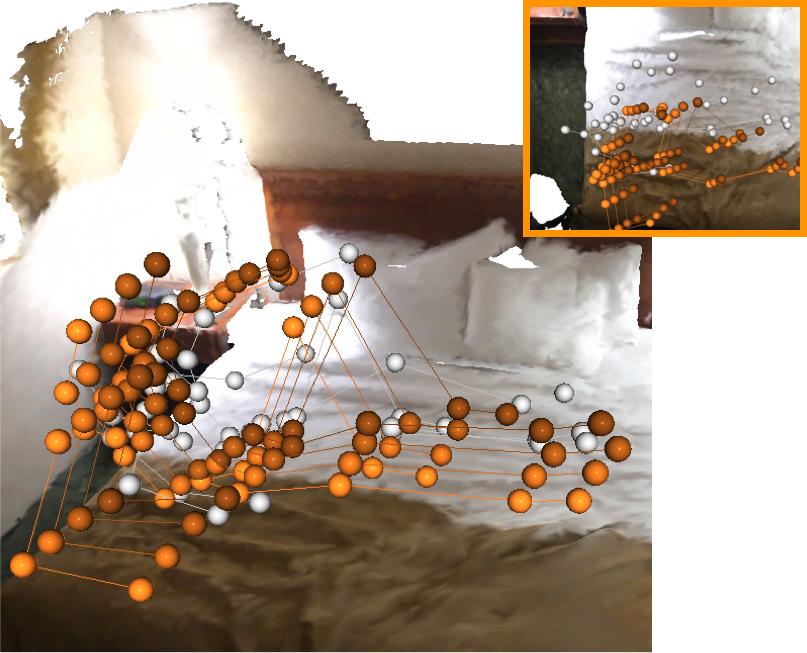}\hspace{-1mm} & \includegraphics[width=0.2\textwidth]{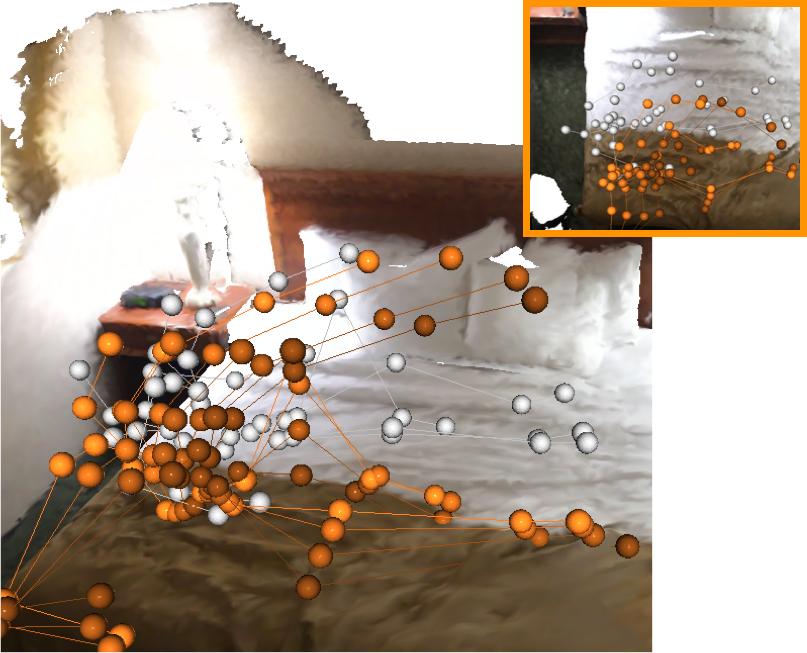}\hspace{-1mm} \\
\end{tabular}    
\caption{This figure compares our method with baseline models on GTA-IM~\cite{cao2020long} (top row), PROX~\cite{hassan2019resolving} (middle row), and HUMANISE~\cite{wang2022humanise} (bottom row). Our method predicts future motion closer to the ground truth.}
\label{fig:qualitative}
\end{figure}

\subsection{Results}
\noindent\textbf{Quantitative results.}
We provide quantitative comparisons across three datasets: GTA-IM~\cite{cao2020long}, PROX~\cite{hassan2019resolving} and HUMANISE~\cite{wang2022humanise} presented in Tables~\ref{tab:gtaim} and \ref{tab:humanise}. Notably, these datasets lack gaze information, with the exception of GIMO~\cite{zheng2022gimo}, for which we report results without gaze, indicated by $^*$. 
Additionally, we have adapted our training strategy to baseline models for a fair comparison. We report the best performance for each model using our training strategy (denoted by $^\dagger$).  Our method consistently outperforms the baseline approaches on all datasets and at all time steps by a substantial margin, both for path and pose predictions. Particularly noteworthy is our method's superiority in path prediction over baseline models on the real-world PROX~\cite{hassan2019resolving} dataset. Our results demonstrate that our model with ground truth mutual distance significantly outperforms baseline models with ground truth contact (w/GT). This superior performance is evident in both global path and local pose metrics, indicating that our method provides a more effective constraint for human motion.
More results on the GIMO~\cite{zheng2022gimo} dataset can be found in the supplementary material.

\begin{figure*}[!ht]
    \centering
    \includegraphics[width=\linewidth]{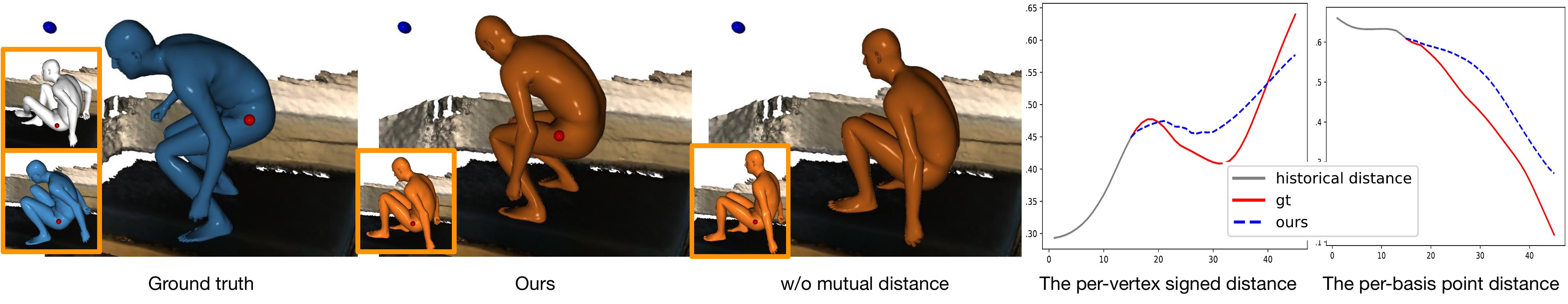}
    \caption{{\bf Ablation of the mutual distance}. 
    The three figures in the first three columns (from left to right) depict
    the ground truth pose for the last frame of future motion, results of our full model, and predictions of our model without mutual distance constraint. The sub-figure in gray is the last observed frame. Other sub-figures depict the predicted middle frame. 
    The blue dot is the scene basis point, and the red dot is the sampled vertex
    on human mesh. The two graphs in the last two columns show the predicted per-vertex
    signed distance and the per-basis point distance for the red and blue point, respectively.
    As shown in the figures, with predicted mutual distance, we can forecast the 'stand-up' action more accurately.     
    }
    \label{fig:distance}
\end{figure*}
\noindent\textbf{Qualitative results.} 
We present qualitative comparisons with baseline methods on GTA-IM~\cite{cao2020long}, HUMANISE~\cite{wang2022humanise}, and PROX~\cite{hassan2019resolving}, in Fig.~\ref{fig:qualitative}. The comparisons further demonstrate that our approach excels in predicting future human motions, aligning closely with ground truth data. In contrast, GIMO~\cite{zheng2022gimo} exhibits issues such as motion freezing and scene penetration. Additionally, we show the predicted human skeletons generated by ContAware~\cite{mao2022contact} and STAG~\cite{scofano2023staged}, which struggle to handle complex motions.

\noindent\textbf{Ablation study.} To gain a more comprehensive understanding, we present ablation results on HUMANISE~\cite{wang2022humanise} dataset in Table \ref{tab:ablation}. We report results for our base models which are defined as our model without scene and mutual distance ('Motion only') and our model without mutual distance ('Scene only'). Notably, there are significant improvements in both path and pose prediction when we incorporate mutual distance into our model. Additionally, we assess the impact of two key components: the proposed per-vertex signed distance ('Ours w/o $\mathbf{B}$') and the per-basis point distance ('Ours w/o $\mathbf{D}$').
Each distance individually contributes to the improvement for both path and pose prediction, with the most significant gains achieved when both distances are utilized. Furthermore, we examine the influence of the global scene representation: point cloud representation ('Ours w/ $P$') and occupancy representation ('Ours w/ $O$'). For point cloud representation, similar to ContAware~\cite{mao2022contact}, we modified the official PVCNN~\cite{liu2019point} code to take point cloud as scene input instead of SDF volume. For occupancy representation, our method uses occupancy volume as scene input. Our model outperforms both the point cloud and occupancy representation across all metrics. We also compare the Basis Point Set (BPS) encoding in PLACE~\cite{zhang2020place} with our per-basis point distance ('Ours w/ $BPS$') by replacing the per-basis point distance with BPS encoding following the official implementation in \cite{zhang2020place}. Moreover, we show results from end-to-end training of our method ('Ours $e2e$'), which encounters a local minimum in the early epochs. Results shown in Fig.~\ref{fig:distance} further demonstrate the effectiveness of our method.


%% file: 06-conclusion.tex
\section{Conclusion}\label{sec:conclu}

In this work, we present an approach for scene-aware human motion forecasting that explicitly models the human-scene interaction. To constrain the whole-body motion, we have proposed a mutual distance representation: the per-vertex signed distance and the per-basis point distance. The mutual distance captures both the detailed local body movements and the global human motion. 
Thanks to the explicit mutual distance constraints, 
Our method generates future human motions that are more plausible and accurate than those produced by state-of-the-art scene-aware motion prediction techniques.

\noindent\textbf{Societal Impacts \& Limitations.} 
The motion prediction results depend on the prediction of the mutual distance is one limitation of our work. Our ablation study using the ground truth mutual distance indicates that enhancing the prediction of mutual distance results in improved motion prediction. This can therefore serve as a potential direction for future research.
In real applications, a potential risk of our model is that without explicit real world physics modeling, it is possible for our model to predict future motions that have imbalanced forces. This may lead to unstable human motion when deploy to real life applications like human-robot interaction such as collision.

\noindent{\textbf{Acknowledgements}} This research was supported in part by the Australia Research Council DECRA Fellowship (DE180100628) and ARC Discovery Grant (DP200102274).